\documentclass[runningheads]{llncs}
\pdfoutput=1
\usepackage{orcidlink}
\usepackage[T1]{fontenc}
\usepackage{lmodern}
\usepackage{graphicx}
\usepackage{hyperref}
\usepackage{color}

\usepackage{amsmath}
\usepackage{amssymb}

\usepackage[textsize=tiny]{todonotes}

\usepackage{rotating} 
\usepackage{booktabs}
\usepackage{eurosym} 
\usepackage{appendix}
\usepackage{algorithm}
\usepackage{algcompatible}
\usepackage{caption}


\newcommand{\xv}{\mathbf{x}}
\newcommand{\Xv}{\bar{\text{\b{\textbf{X}}}}}
\newcommand{\Xspace}{\mathcal{X}}
\newcommand{\R}{\mathbb{R}}
\newcommand{\fh}{\hat{f}}

\newcommand{\Prob}{I\kern-0.15em P} 
\newcommand{\I}{\mathbb{I}} 
\newcommand{\Xeval}{\mathbf{E}}
\newcommand{\Bbar}{\bar{\text{\b{B}}}}

\newcommand{\citep}[1]{\cite{#1}}

\begin{document}
\title{Interpretable Regional Descriptors: Hyperbox-Based Local Explanations}
%
\titlerunning{Interpretable Regional Descriptors}
%
\author{Susanne Dandl\inst{1,2}~\orcidlink{0000-0003-4324-4163}  \and
 Giuseppe Casalicchio\inst{1,2}~\orcidlink{0000-0001-5324-5966} \and
 Bernd Bischl\inst{1,2}~\orcidlink{0000-0001-6002-6980} \and 
 Ludwig Bothmann\inst{1,2}~\orcidlink{0000-0002-1471-6582}}
\authorrunning{S. Dandl et al.}

 \institute{Department of Statistics, LMU Munich, Ludwigstr. 33, 80539 Munich, Germany \and Munich Center for Machine Learning (MCML), Munich, Germany \\
\email{Ludwig.Bothmann@stat.uni-muenchen.de}}
\maketitle              
\begin{abstract}

This work introduces interpretable regional descriptors, or IRDs, for local, model-agnostic interpretations. IRDs are hyperboxes that describe how an observation's feature values can be changed without affecting its prediction.
They justify a prediction by providing a set of ``even if'' arguments (semi-factual explanations), and they indicate which features affect a prediction and whether pointwise biases or implausibilities exist. A concrete use case shows that this is valuable for both machine learning modelers and persons subject to a decision.
We formalize the search for IRDs as an optimization problem and introduce a unifying framework for computing IRDs that covers desiderata, initialization techniques, and a post-processing method. We show how existing hyperbox methods can be adapted to fit into this unified framework. 
A benchmark study compares the methods based on several quality measures and identifies two strategies to improve IRDs. 

\end{abstract}

\section{Introduction}
\label{sec:intro}

Supervised machine learning (ML) models are widely used due to their good predictive performance, but they are often difficult to interpret due to their complexity.
Post-hoc interpretation methods from the field of interpretable machine learning (IML) can help to draw conclusions about the inner processes of these models.
Two types of interpretation methods can be differentiated: local methods that explain individual predictions, and global methods that explain the expected behavior of the model in general. 
Doshi-Velez and Kim \cite{doshi_etal_iml_2017} define model interpretability as ``the ability to explain or to present in understandable terms to a human''. 
A topological form that satisfies this notion of interpretability is a hyperbox. 
In this work, we investigate hyperboxes as local interpretations that describe how the feature values of an observation can be changed without affecting its prediction. 
We call these boxes interpretable regional descriptors (IRDs). 
IRDs describe feature spaces by intervals for real-valued features and subsets of all possible classes for categorical features (see Table~\ref{tab:credit}).

\subsection{Motivating Example for the Use of IRDs}
Consider bank lending as a motivating example: a customer applies for a credit of \euro4000 at a bank to buy a new car. She is 22 years old, skilled, lives in a rented accommodation, has few savings and a moderate balance on her checking account. An ML model predicts whether the credit is of low, moderate or high risk. Due to a moderate risk prediction, the bank rejects the application.
The IRD in Table~\ref{tab:credit} answers the question ``to what extent the feature or multiple features can be changed such that the prediction is still in the moderate risk class''. From an IRD, multiple insights into a prediction can be obtained.

\begin{table}[t!]
	\caption{Example based on the credit dataset \cite{uci,kaggle16} with 9 \textit{features}. The second column shows the values of a  \textit{customer} with a moderate credit risk prediction. The \textit{IRD} (generated by MaxBox \& post-processing (Section~\ref{sec:generation})) shows how all features could be changed simultaneously so that the credit is still of moderate risk. The \textit{1-dim IRD} shows how a single feature could be changed without changing the prediction (keeping the other features fixed). For features in the upper half, the IRD covers the full observed value \textit{range} in the training data.} 
	\label{tab:credit}
	\footnotesize
	\centering
	
	\begin{tabular}{lllll}
		\hline
		Feature & Customer & IRD & 1-dim IRD & Range \\ 
		\hline
		sex & female & \{female, male\} & \{female, male\} & \{female, male\} \\ 
		saving.accounts & little & \{little, moderate & \{little, moderate, & \{little, moderate, \\ 
		& & rich\} & rich\} & rich\} \\
		purpose & car & \{car, radio/TV, & \{car, radio/TV, & \{car, radio/TV,  \\ 
		& & furniture, others\} & furniture, others\} & furniture, others\} \\
		\hline
		age & 22 & [19, 22] & [19, 75] & [19, 75] \\ 
		job & skilled & \{skilled, highly &  \{unskilled, skilled, &  \{unskilled, skilled, \\
		& & skilled\} & highly skilled\} & highly skilled\} \\
		housing & rent & \{rent\} & \{own, free, rent\} & \{own, free, rent\} \\ 
		checking.account & moderate & \{little, moderate\} & \{little, moderate\} & \{little, moderate, \\
		& & & & rich\} \\
		credit.amount & 4000 & [4000, 5389] & [2127, 8424] & [276, 18424] \\ 
		duration & 30 & [26, 33] & [6, 44] & [6, 72] \\ 
		\hline
	\end{tabular}
\end{table}

First, IRDs offer a set of semi-factual explanations (SFEs) –- also called a fortiori arguments –- to justify a decision in the form of “even if” statements \cite{nugent_gaining_2009}.
For these statements to be convincing, domain knowledge is required, e.g., that higher balances in the savings and checking account, and that higher skilled jobs decrease the risk for a bank. Given such knowledge, a multitude of semi-factual explanations can be derived from the IRD of Table~\ref{tab:credit} that (1) justify that a person is in the moderate risk class instead of the low risk class (e.g., ``even if you had a moderate balance in the savings account and become highly skilled, your credit is still of moderate risk''), and that (2) justify that a person is in the moderate risk class instead of the high risk class (``even if you only have little balance in your checking account, your credit would still be of moderate risk''). The latter statement also reveals a ``safety bound'' if some of the features change in the undesirable direction (high risk class) in the future. 

Second, the interval width or cardinality of a feature in an IRD relative to its entire feature space can indicate whether a feature affects a prediction locally (if Theorems~\ref{theorem:sparsity}~and~\ref{theorem:detectionpower} hold). 
For example, compared to the credit amount or
balance status of her checking account, savings or purpose seem
to have no local effect on the prediction in the bank lending
example, since the regional descriptor covers the whole observed feature range in these two dimensions. These insights also reveal what can be options to change a given prediction.\footnote{However, the concrete strategies can only reveal counterfactual explanations \cite{wachter_cfexp_2018}.}

Third, IRDs are tools for model auditing. If the insights from a box (e.g., a semi-factual explanation) agree with domain knowledge, users have more trust in the model, while disagreement helps to reveal unintended pointwise biases or implausibilities of a model.
For example, an IRD that does not cover male customers \textit{might} indicate that the model classifies individuals differently based on gender.\footnote{Note that if all genders are part of the box, it does not mean the model is fair.}
An IRD that covers a credit amount of \euro300 and high balances in the checking account could be an indicator of an inaccurate model because such customers should pose only a low risk to the bank.
In addition to credit risk, we show other practical applications of IRDs in Appendix~A. 

\subsection{Contributions}
Our contributions are: 1) We introduce IRDs as a new class of local interpretations to describe regions in the feature space that do not affect the prediction of an observation; 2) We formalize the search for IRDs as an optimization problem and develop desired properties of IRD methods; 3) We introduce a unifying framework for computing IRDs including initialization techniques and post-processing methods; 4) We show how existing hyperbox methods from data mining or IML can be adapted to fit into our unified framework; 5) We present a set of quality measures and compare our derived methods accordingly in a benchmark study; 
6) We provide open-access repositories with an R package for the implemented approaches and the code for replicating the benchmark study.\footnote{Links will be shared upon acceptance. For review, we attached a zip file.} 

\section{Methodology}
\label{sec:methodology}

Let $\fh:\Xspace \rightarrow \R$ be the prediction function of an ML model, where $\Xspace$ denotes a $p$ dimensional feature space. For classification models, we consider a pre-defined class of interest for which $\fh$ returns the predicted score or probability.

\subsection{Formalizing the General Task for IRDs}
\label{subsec:task}

Our goal is to find the largest hyperbox $B$ covering a point of interest $\xv' \in \Xspace$ where all data points in $B$ have a sufficiently close prediction to $\fh(\xv')$.
The hyperbox $B$ should have $p$ dimensions $B = B_1 \times ... \times B_p$
\begin{equation}
\text{ with } B_j = \begin{cases}
\{c|c \in \Xspace_j\} & \textrm{categorical } X_j \\
[l_j, u_j] \subseteq \Xspace_j & \, \textrm{numeric } X_j \\
\end{cases},
\label{eq:Bj}
\end{equation}
consisting of intervals for numeric features and a subset of possible classes for categorical features.
$\Xspace_j$ reflects the value space of the $j$th feature $X_j$.    
In accordance with Lemhadri et al. \cite{lemhadri_li_hastie_2022_rbx}, a prediction is sufficiently close if it falls into a \textit{closeness region}, which is a user-defined prediction interval $Y' = [\fh(\xv') - \epsilon_L, \fh(\xv') + \epsilon_H]$ with $\epsilon_L, \epsilon_H \in \R_{\ge 0}$.\footnote{For classification models, $Y' \subset [0, 1]$ must hold.}
In the bank lending example, the closeness region should cover all model predictions that lead to the moderate risk class, e.g., a predicted probability of 30-60 \% of defaulting, i.e., $Y' = [0.3, 0.6]$. To operationalize the above goal, we need three measures \cite{ribeiro_anchors_2018,sharma_maire_2021}: 
\begin{enumerate}
	\itemsep1em
	\item 
	%
	$
	coverage(B) = \mathbb{P}(\xv \in B|\xv \in \Xspace),
	$
	which measures how much a hyperbox covers the entire feature space. Since, in practice, not all $\xv \in \Xspace$ are observable, we use an empirical approximation given data $(\xv_i)_{1\le i \le n}$ with $\xv_i \in \Xspace$
	\begin{equation}
	\widehat{coverage}(B) = \frac{1}{n} \sum_{i = 1}^n \I(\xv_i \in B).
	\label{eq:empcoverage}
	\end{equation}
	\item 
	$
	precision(B) = \mathbb{P}(\fh(\xv) \in Y'|\xv \in B),
	$
	the fraction of points within a box $B$ whose predictions are inside $Y'$. Again, we use an empirical approximation
	\begin{equation}
	\widehat{precision}(B) = \frac{\sum_{i = 1}^n \I(\xv_i \in B \wedge f(\xv_i) \in Y')}{\sum_{i = 1}^n \I(\xv_i \in B)}.
	\label{eq:empprecision}
	\end{equation}
	
	\item an indicator of whether $B$ covers $\xv'$
	\begin{equation}
	locality(B) = \I(\xv' \in B).  
	\label{eq:locality}
	\end{equation}
\end{enumerate}
The following operationalizes the search for an IRD \cite{ribeiro_anchors_2018}:\footnote{For this, we extended the optimization task of Ribeiro et al.\ \cite{ribeiro_anchors_2018} to target IRDs by aiming for a precision of 1 and by including the locality constraint.} 
\begin{equation}
\begin{split}
&\underset{B \subseteq \Xspace}{arg\,max}(\widehat{coverage}(B)) \\
& \text{ s.t. } \widehat{precision}(B) = 1 \text{ and } locality(B) = 1.
\end{split}
\label{eq:optimtask}
\end{equation} 
\begin{definition}
	A box is maximal if and only if  no box could be added under full precision, such that for all numeric $X_j$, it holds that $(\nexists\, x_j \in \Xspace_j \wedge x_j < l_j: precision(B \cup [x_j, l_j]) = 1) \wedge (\nexists\, x_j \in \Xspace_j \wedge x_j > u_j: precision(B \cup [u_j, x_j]) = 1)$, and for all categorical $X_j$, it holds that
	$(\nexists\, x_j \in \Xspace_j\setminus B_j : precision(B \cup x_j) = 1)$.
	\label{def:maximality}
\end{definition}
A box $B$ with maximum coverage satisfies this maximality property.
We aim for a maximal $B$, since $B$ can then detect features that are not locally relevant for a prediction $\fh(\xv')$. We prove the following in Appendix~B.
\begin{theorem}
	If $B$ is maximal, $B_j = [min(\Xspace_j), max(\Xspace_j)]$ holds for a feature $X_j$ that is not involved in the model $\fh$.
	\label{theorem:sparsity}
\end{theorem}
Similarly, we aim for homogeneous boxes $B$ such that $precision(B) = 1$. Then, $B$ can detect features that are locally relevant for $\fh(\xv')$.
We prove the following in Appendix~C. 
\begin{theorem}
	If $precision(B) = 1$, $B_j \subset \Xspace_j$ holds for a feature that is locally relevant for $\fh(\xv')$.
	\label{theorem:detectionpower}
\end{theorem}

\subsection{Desiderata for IRDs}
\label{subsec:desiderata}

In Section~\ref{sec:relatedwork}, we discuss related methods to generate $B$.
The suitability of these methods as IRD methods relies on whether they consider all objectives of Eq.~(\ref{eq:optimtask}) and whether they satisfy the following desired properties for IRDs.

\paragraph{Interpretability}
In order for $B$ to be interpretable, we only consider methods that return a \textit{single} $p$-dimensional hyperbox.
The hyperrectangular structure of $B$ allows for a natural interpretation, which is not the case for hyperellipsoids or polytopes formed by halfspaces \cite{lemhadri_li_hastie_2022_rbx}. 
According to Eq.~(\ref{eq:optimtask}), $B$ needs to cover $\xv'$, which is the case if the following holds:
$\forall j \in \{1, ..., p\}: x_j' \in B_j$.

\paragraph{Model-agnosticism}

The definition of $\fh$ does not pose any restrictions on the ML model or the feature space. Therefore, methods should be model-agnostic such that they could explain both regression or classification models with various feature types (binary, nominal, ordinal or continuous). 

\paragraph{Sparsity constraints}
Eckstein et al.\ \cite{eckstein_maxbox_2002} proved that the optimization task for the maximum box problem is $\mathcal{NP}$-hard if the features defining the box are not fixed. 
This also applies to the search for IRDs, which only additionally requires $\xv' \in B$.
Since the search space for hyperboxes grows with the number of features, it is infeasible to consider all potential solutions.
Furthermore, the fact that IRDs have as many dimensions as the dataset impedes their interpretability – the very goal of IRDs in the first place.
To reduce the number of features, methods should be able to adhere to user-defined sparsity constraints such that for some features $X_j$, $B_j = x_j'$.
Section~\ref{sec:conclusion} discusses other solutions.

\section{Related Work}
\label{sec:relatedwork}

\begin{table}[t]
	\centering
	\footnotesize
	\caption{Overview of approaches that search for hyperboxes in feature spaces.}
	\label{tab:relatedwork}
	\begin{tabular}{l|ccc|ccc}
		& \multicolumn{3}{c|}{\textbf{Objectives}}  & \multicolumn{3}{c}{\textbf{Desiderata}} \\
		& Coverage & Precision &  Locality & Interpretable & Agnostic & Sparse  \\
		\hline
		\textbf{Level set methods} & & & & &  \\
		PBnB \cite{zabinsky_partition-based_2020,zabinsky_adaptive_2011} & $\surd$ & $\surd$ & $\times$ & $\times$ & $\surd$ & $\times$ \\
		\hline
		\textbf{Data mining} & & & & & \\
		MaxBox \cite{eckstein_maxbox_2002} & $\surd$ & $\surd$ & $\times$ & $\surd$ & $\times$ & $\times$ \\
		PRIM \cite{friedman_bump_1999} & $\times$ & $\times$ & $\times$ & $\surd$ & $\times$ & $\times$ \\
		\hline
		\textbf{Post-hoc IML}  & & & & & \\
		Anchors \cite{ribeiro_anchors_2018} & $\surd$ & $\surd$ & $\surd$ & $\surd$ & $\times$ & $\times$ \\ 
		MAIRE \cite{sharma_maire_2021} & $\surd$ & $\surd$ & $\surd$ & $\surd$ & $\times$ & $\times$ \\
		LORE \cite{guidotti_factual_2019,guidotti_factual_2022} & $\times$ & $\times$  & $\surd$ & $\surd$ & $\surd$ & $\times$ \\
		\hline
		\textbf{Interpretable classifier} & & & & & \\
		Column generation \cite{dash_boolean_2018} &  $\surd$ & $\surd$ & $\times$ & $\times$ & $\surd$ & $\times$ \\
	\end{tabular}
\end{table}

The optimization task of Eq.~(\ref{eq:optimtask}) can be understood mathematically as finding the preimage of prediction values $\in Y'$ in the neighborhood of $\xv'$. 
Therefore, IRDs can be seen as a subset of a level set for function values $\in Y'$. Level set approximations often consist of points  \cite{emmerich_quality_2013}, and only a few approaches approximate these via hyperboxes \cite{zabinsky_partition-based_2020,zabinsky_adaptive_2011}. These methods produce multiple boxes instead of a single one and do not require to contain a given $\xv'$.
Hence, they are not interpretable in our sense and, therefore, not useful to produce IRDs.

In data mining, \cite{eckstein_maxbox_2002} proposed a maximum box (MaxBox) approach for datasets with binary outcomes to find the largest homogeneous hyperbox w.r.t.\ the positive class.
Friedman and Fisher \cite{friedman_bump_1999} derived the Patient Rule Induction Method (PRIM) for seeking
boxes in the feature space in which the outcome mean is high.  
Both approaches do not require $\xv'$ to be in the box. 

As described earlier, IRDs may also be seen as a method to summarize a multitude of SFEs.
Most proposed methods for SFEs return only a single point as an explanation \cite{dhurandhar_pertinent_2018,kenny_etal_semifactualsindl_2021,nugent_gaining_2009}.
In contrast, the approach by Guidotti et al.\ \cite{guidotti_factual_2019,guidotti_factual_2022} returns a set of SFEs using surrogate trees.
Their approach reveals which feature values are most important for deriving a prediction by following the path to the point of interest. 
The reliability of surrogate trees depends on the assumption that the tree can adequately replicate the underlying model, which is often not the case.
Furthermore, IRDs require homogeneous boxes, which is only possible with overfitting/deep-grown trees.
Therefore, the tree structure is only suitable for deriving SFEs when the underlying model is tree-based \cite{fernandez_factual_2022,stepin_generation_2020}. 

An IML method that utilizes hyperboxes is the Anchors approach \cite{ribeiro_anchors_2018}.
The returned hyperbox indicates how features must be fixed or anchored to prevent a model from changing the classification of a data point.
Anchors were originally proposed to aim for hyperboxes that also partly cover observations of other classes; a precision of $0.95$ is the default in its implementation \cite{ribeiro_anchorsgithub_2022}.
Although the precision can be changed to $1$, Anchors are nevertheless not suitable for the generation of IRDs due to their limited search space: Either the box boundary of a feature is set to the full feature range observed in the data, or to the value of $\xv$. This bears the risk of ``overly specific anchors'' with low coverage \cite{ribeiro_anchors_2018}.
To generate boxes with larger coverage, features can be binned beforehand. However, no established discretization technique for Anchors exists so far and the optimization procedure underlying Anchors does not allow adaptions of the bins during optimization.

To overcome the discretization problem, Sharma et al.\ \cite{sharma_maire_2021} proposed the model-agnostic interpretable rule extraction (MAIRE) procedure.
MAIRE finds more optimal boundaries for continuous features via gradient-based optimization.
It still does not allow a more precise choice for categorical features; either the box allows no changes to a feature or it covers all possible values of a feature. 

Dash et al.\ \cite{dash_boolean_2018} proposed a classifier based on a set of hyperboxes. The method focuses on an optimal combination of hyperboxes to derive an accurate model for inputs from the whole feature space using column generation.
As such, the method does not focus on locality and is not interpretable in our sense.

Table~\ref{tab:relatedwork} summarizes whether the addressed methods are suitable for generating IRDs. Overall, none of the methods satisfies all objectives of Eq.~(\ref{eq:optimtask}) and desiderata from Section \ref{subsec:desiderata}. Specifically, none of them addresses sparsity constraints, and only a few are model-agnostic. 
In Section \ref{subsec:irdmethods}, we modify MaxBox, PRIM, and MAIRE such that they fulfill all of our requirements to transform them into useful IRD methods.
All other methods cannot be modified to the required extent, since their underlying, irreplaceable optimization methods either target multiple boxes or different search spaces.
The latter applies in particular to Anchors. However, the method serves as a baseline method for our benchmark study in Section~\ref{sec:evaluation}. 

\section{Generating IRDs}
\label{sec:generation}

We now present a unifying framework for generating IRDs, which consists of
four steps: restriction, selection, initialization, and optimization. 
Optionally, a post-processing step can be conducted (Section~\ref{subsec:postprocessing}).

\subsection{Restriction of the Search Space}
\label{subsec:searchspace}

To restrict the initial search space for $B$, we propose a simple procedure to find the largest local box $\Bbar$ of $\xv'$ such that $B \subset \Bbar$.
For a continuous feature $X_j$, we vary its value $x'_j$ of $\xv'$ on an equidistant grid. Upper and lower bounds of $\Bbar_j$ are set to the minimal changes in $x'_j$, yielding a prediction outside $Y'$. This approach is similar to individual conditional expectation (ICE) values \cite{goldstein_etal_ice_2015}. 
For a categorical feature $X_j$, $\Bbar_j$ comprises all classes of $\Xspace_j$ that still lead to a prediction $\in Y'$ after adapting $x'_j$ of $\xv'$. 
If a user sets the sparsity constraint that feature $X_j$ is immutable, $\Bbar_j = x_j'$ must hold. We prove the following in Appendix~D.
\begin{theorem}
	For any box $B$ that solves the optimization problem of Eq.~(\ref{eq:optimtask}) it holds that $B \subseteq \Bbar$.
	\label{theorem:barBbar}
\end{theorem} 

\subsection{Selection of the Underlying Dataset}
All methods need a dataset $\Xv$ consisting of $\xv \in \Xspace$ as an input. This dataset is used for evaluating (competing) boxes w.r.t.\ the empirical versions of coverage and precision (Eq.~(\ref{eq:empcoverage}) and Eq.~(\ref{eq:empprecision})). For some methods, the dataset also offers a set of potential box boundaries to be evaluated.
A suitable dataset is the training data. Since only instances $\in \Bbar$ are relevant (Theorem~\ref{theorem:barBbar}), we remove all instances $\not \in \Bbar$ from $\Xv$. Consequently, $x_j = x_j' \, \forall \, \xv \in \Xv$ holds for all immutable features $X_j$.
More features and sparsity constraints increase the risk that $\Xv$ is only sparsely populated around $\xv'$.
Since we aim for IRDs that are faithful to the model and not to the data-generating process (DGP), new data can be generated by uniformly sampling from the admissible feature ranges of $\Bbar$. 
In Section~\ref{sec:evaluation}, we inspect how double-in-size sampled data within $\Bbar$ \footnote{Double-in-size refers to the size of the training data, not of $\Xv$.} affects the quality of IRDs and IRD methods compared to using training data.

\subsection{Initialization of a Box}
All methods require an initial box $B$ as an input, which is either set to the largest local box $\Bbar$ covering all $\Xv$ or the smallest box possible which only contains $\xv'$. 
We define methods that start with the largest local box as top-down IRD methods, and methods that start with the smallest box possible as bottom-up methods.



\subsection{Optimization of Box Boundaries}
\label{subsec:irdmethods}
The last step comprises the optimization of the box boundaries. 
Top-down methods iteratively shrink the box boundaries of the largest local box to improve the box's precision (upholding that $\xv' \in B$), while bottom-up methods iteratively enlarge the box boundaries of the smallest box to improve the box's coverage (upholding the precision at 1).
In this section, we describe the MaxBox, MAIRE, and PRIM approaches and our extensions such that the methods optimize Eq.~(\ref{eq:optimtask}) and fulfill the desiderata of Section~\ref{subsec:desiderata}. Pseudocodes and illustrations of the inner workings of the extended approaches are given in Appendix~E. All methods receive as input a dataset $\Xv$ and an initial box $B$.

\paragraph{MaxBox -- Top-down Method}
MaxBox was originally proposed for binary classification problems -- with a positive and negative class.
The method starts with the largest box covering all data.
A branch and bound (BnB) algorithm \cite{land_doig_1960_bandb} inspects the options to shrink the box to optimize its precision w.r.t.\ the positive class.
The branching rule creates new boxes by bracketing out a sample $\xv$ of the negative class, such that the box is shrunk to be either below or above the values of $\xv$ in at least one feature dimension (categorical features are one-hot encoded). Estimates of the upper bound for the coverage of a box determine which imprecise box is branched next, which sample is used for branching, and which boxes are discarded because their upper bound does not exceed the coverage of the current largest homogeneous box. If no boxes to shrink are left, the largest homogeneous box is returned as an IRD.

\textit{Extensions} By labeling observations with predictions $ \in Y'$ as positive, the approach becomes model-agnostic. 
Since the original algorithm does not consider whether corresponding boxes still include $\xv'$, we adapted the approach to discard boxes that do not contain $\xv'$ to guarantee locality.

\paragraph{PRIM -- Top-down Method}
The method originally aims for boxes with a high average outcome. The procedure starts with a box that includes all points. In the peeling phase, PRIM iteratively identifies a set of eligible subboxes (defined by the $\alpha$- and (1-$\alpha$)-quantile for numeric features and each present category for categorical features) and peels off the subbox that results in the highest average outcome after exclusion.
This step is repeated until the number of points included in the box drops below a fraction of the total number of points.
In the pasting phase, the box is iteratively enlarged by adding the subbox that increases the outcome mean the most. These subboxes consist of at least $\alpha$ observations with the nearest lower or higher values in one dimension (numeric $X_j$) or with a new category (categorical $X_j$). 

\textit{Extensions} We adapted the approach to target Eq.~(\ref{eq:optimtask}): in each peeling iteration, the subbox is excluded such that the resulting box has the highest precision (coverage acts as a tiebreaker), and in each pasting iteration, the largest homogeneous subbox is added. If the precision and coverage are not sufficient to select a best box for peeling or pasting, a subbox is randomly selected from the best ones.
Peeling stops as soon as the resulting box is homogeneous, while pasting stops as soon as there exists no homogeneous box to add. 
Furthermore, only subboxes that do not cover $\xv'$ are peeled.
According to the authors' recommendation, we use $\alpha = 0.05$ for the benchmark study (Section~\ref{sec:evaluation}).

\paragraph{MAIRE -- Bottom-up Method} 
The method starts with a box covering $\xv'$. 
In each iteration, the box boundaries are adapted via ADAM \cite{kigma_2017_adam} by optimizing a differentiable approximation of the coverage measure. If the precision falls below a certain threshold or $\xv'$ is not part of the box, the method additionally optimizes a differentiable version of Eq.~(\ref{eq:empprecision}) and Eq.~(\ref{eq:locality}), respectively.
MAIRE stops after a specified number of iterations. In the end, the method returns the largest homogeneous box over the iterations.

\textit{Extensions} The method requires 0-1-scaled features. To overcome the one-vs-all issue for categorical features (Section~\ref{sec:relatedwork}), we one-hot-encode categorical features.
We implemented a convergence criterion for a fair comparison with the other (convergent) approaches: we let MAIRE enlarge the box boundaries until the precision falls below $1$, then MAIRE is only allowed to run for another $100$ iterations.
The implementation for the experiments in Section~\ref{sec:evaluation} is based on the authors' implementation \cite{sharma_maire_2021} with the discussed modifications. 
The hyperparameters were set according to the authors' recommendations. We only set the precision threshold to 1, rather than 0.95.

\subsection{Post-processing}
\label{subsec:postprocessing}
All methods described in the previous section determine box boundaries based on a finite number of data points in $\Xv$.
The limited access carries the risk that some regions of the feature space are not represented in $\Xv$ and that the boundaries of a generated $B$ are suboptimal: There could be areas in $B$ that have predictions $\notin Y'$, or there could be adjacent areas outside of $B$ that also have predictions $\in Y'$. 
To improve the box boundaries of a given box $B$, we developed the following post-processing method using newly sampled data. 
The procedure consists of peeling and pasting as PRIM.

First, the precision of $B$ is measured based on newly sampled data. If $\exists \xv \in B$ with $\fh(\xv) \notin Y'$, subboxes with the lowest precision in proportion to their size (according to newly sampled data within this subbox) are iteratively peeled. If all subboxes to peel are homogeneous, peeling stops. In the subsequent pasting step, the largest subboxes that proved to be homogeneous (according to newly sampled data within this subbox) are added. If the best box cannot be clearly determined (because several boxes have the same precision and coverage), a subbox is randomly chosen.
The method has three hyperparameters: the number of samples used for evaluation, the relative box size (in relation to the size of $\mathcal{X}_j$) for peeling or pasting boxes for continuous features, and a threshold for the minimum box size.
The latter acts as a stopping criterion for pasting. If no homogeneous subbox can be added, the relative box size to add for continuous features is halved as long as the relative box size is not lower than the threshold.
The pseudocode of our method displays Appendix~F.

Section~\ref{sec:evaluation} investigates whether our post-processing method improves IRDs. For the experiments, we set the number of samples to evaluate boxes to 100, the relative box size to 0.1, and the threshold for the minimum box size to 0.05. 

\section{Quality Measures}
\label{sec:quality}

We now present a set of quality measures for \textit{generated IRDs} and \textit{IRD methods}. These measures apply to a single instance $\xv'$ to be explained, where $B$ is the returned IRD of $\xv'$ of an IRD method $G$.
The assessment requires evaluation data $\Xeval$ consisting of $\xv \in \Xspace$; 
for the benchmark study in Section~\ref{sec:evaluation}, we use training data and new data uniformly sampled from $\Bbar$. 
Training data helps to assess whether the methods use the training data appropriately 
during IRD generation (e.g., precision should be 1), while a proliferated number of newly generated data $\in \Bbar$ leads to a more precise evaluation w.r.t.\ the model, not the DGP. 


\paragraph{Locality}
The IRD should cover $\xv'$. This property is fulfilled if $locality(B) = \I(\xv' \in B)$ equals 1.

\paragraph{Coverage} Given two IRDs with equal precision, we prefer the one with higher coverage (Eq.~(\ref{eq:empcoverage})).
To evaluate the coverage, we use samples $\xv \in \Xeval$ from the connected convex level set $\mathcal{L}$ covering $\xv'$.
\begin{definition}
	A datapoint $\xv$ with $\fh(\xv) \in Y'$ is part of $\mathcal{L}$ of $\xv'$ iff there exists a path between $\xv$ and $\xv'$ for which all intermediate points have a prediction $\in Y'$.
\end{definition}
Paths are identified via the identification algorithm of Kuratomi et al.\ \cite{kuratomi_juice_2022}, details are given in Appendix~G.

\paragraph{Precision} Given two IRDs with equal coverage, the IRD with higher precision is preferred (Eq.~(\ref{eq:empprecision})).

\paragraph{Maximality} A box should be maximal according to Definition~\ref{def:maximality} based on $\xv \in \Xeval$ instead of $\xv \in \Xspace$.


\paragraph{No. of Calls} Lower number of calls to $\fh$ of an IRD method are preferred.


\paragraph{Robustness} If we rerun method $G$ on the same $\xv'$ and $\fh$ $R$ times using the same $\Xv$, the produced IRDs $B_1, ..., B_R$ should overlap with the originally produced $B$, 
such that
$robustness(G) = \underset{k \in \{1, ..., R\}}{\operatorname{min}} \frac{\sum_{\xv \in \Xeval} \I(\xv \in B \cap B_k)}{\sum_{\xv \in \Xeval} \I(\xv \in B \cup B_k)} $ has a high value.


\section{Performance Evaluation}
\label{sec:evaluation}

In a benchmark study, we address the following research questions (RQs): 

\begin{enumerate}
	\item Based on the stated quality measures of Section~\ref{sec:quality}, how do the different methods of Section~\ref{subsec:irdmethods} perform against each other and the baseline method when training data are used as $\Xv$ (without post-processing)?
	
	\item What effect do double-in-size sampled data originating from $\Bbar$ have on the quality of the IRDs and methods compared to using training data? 
	
	\item What effect does the post-processing (Section~\ref{subsec:postprocessing}) have on the quality of the IRD methods?
\end{enumerate}
As a baseline method, we use the Anchors approach \cite{ribeiro_anchors_2018} with a precision of 1 and 20-quantile-based bins for numeric features (see Section~\ref{sec:relatedwork} for details).

\subsection{Setup}

To answer the RQs, we utilize six datasets available on the OpenML platform \cite{vanschoren_etal_openml_2014}, either with a binary, multi-class or continuous target variable. 
Table~\ref{tab:data} summarizes the datasets' dimensions as well as the target and feature types. 
Before training a model, five randomly sampled datapoints were excluded from the datasets to be $\xv'$.
On each of the datasets, four models are trained: a hyperbox model, a logistic regression/multinomial/linear model (depending on the outcome), a neural network with one hidden layer, and a random forest model.
The number of trees for the random forest model and the neurons on the hidden layer are tuned (details are given in Appendix~H).
The hyperbox model is derived from a classification and regression tree (CART) model for each $\xv'$ individually. For a given $\xv'$, the post-processed model predicts 1 if a point falls in the same terminal node as $\xv'$ and 0 otherwise.\footnote{
	The true hyperbox of the CART model might be larger than the terminal node-induced hyperbox (see Figure~6 in the Appendix).}

For classification models, the prediction function returns the probability of the class with the highest probability for $\xv$. For binary targets, we set $Y' = [0.5, 1]$.
For regression and multi-class targets, $Y'$ is set to $[\fh(\xv) - \delta, \fh(\xv) + \delta]$ with $\delta$ as the standard deviation of predictions $\fh$ of the training data. For multi-class, the interval is additionally capped between $0$ and $1$.
For each dataset, model, and $\xv'$, we generate IRDs with MaxBox, PRIM, and MAIRE, as well as Anchors -- our baseline method. The hyperparameters of the methods were set according to Section~\ref{sec:generation}.
The methods were either run on training or on uniformly sampled data from $\Bbar$ (RQ 2), and either without or with post-processing (RQ 3).
For the robustness evaluation, we repeated the experiments $R$ = 5 times.

The methods and their generated IRDs were evaluated based on the performance measures of Section~\ref{sec:evaluation} -- either evaluated on the training data or $1000$ new instances sampled uniformly from $\Bbar$.
We also compared the methods statistically by conducting Wilcoxon rank-sum tests for the hypothesis that the distribution of the coverage and precision values do not differ between two (IRD) methods (RQ 1), for a method using training vs. sampled data (RQ 2), and for a method without vs. with post-processing (RQ 3).
The experiments were conducted on a computer with a 2.60 GHz Intel(R) Xeon(R) processor, and 32 CPUs. Overall, generating the boxes took 63 hours spread over 20 CPUs. The five repetitions for the robustness evaluation required another 316 hours. 

\begin{table}[t]
	\centering
	\footnotesize
	\caption{\label{tab:data} Overview of benchmark datasets. ID: OpenML id; Type: target type; Obs: number of rows; Cont/Cat: number of continuous/categorical features.}
	\begin{tabular}{llllllp{7.4cm}}
		\hline
		Name & ID & Type & Obs & Cont & Cat \\ \hline
		diabetes & 37 & binary & 768 & 8 & 0\\
		tic\_tac\_toe & 50 & binary & 958 & 0 & 9\\
		\hline
		cmc & 23 & three-class & 1473 & 2 & 7 \\
		vehicle & 54 & four-class  & 846 & 18 & 0 \\
		\hline
		no2 & 886 & regression & 500 & 7 &  0\\
		plasma\_retinol & 511 & regression  & 315 & 10 & 3 \\
		\hline
	\end{tabular}
\end{table}

\begin{table}[t]
	\centering
	\footnotesize
	\begingroup\footnotesize
	\caption{\label{tab:bencheval} Comparison of methods  w.r.t.\ maximality and no.\ of calls to $\fh$ averaged over all datasets, models and $\xv'$. Each method was run or evaluated on training data or uniformly sampled data from $\Bbar$, and without (0) or with (1) post-processing. Higher maximality and lower no.\ of calls are better.}
	\begin{tabular}{l|rr|rr|rr|rr|rr|rr}
		\toprule 
		& \multicolumn{6}{c|}{\textbf{Traindata}} & \multicolumn{6}{c}{\textbf{Sampled}} \\
		& \multicolumn{2}{c}{Max$_{train}$} & \multicolumn{2}{c}{Max$_{samp}$} & \multicolumn{2}{c|}{No.\ calls to $\fh$} & \multicolumn{2}{c}{Max$_{train}$} & \multicolumn{2}{c}{Max$_{samp}$} & \multicolumn{2}{c}{No.\ calls to $\fh$} \\
		\cmidrule(l){2-3} \cmidrule(l){4-5} \cmidrule(l){6-7} \cmidrule(l){8-9} \cmidrule(l){10-11} \cmidrule(l){12-13}   
		& 0 & 1 & 0 & 1 & 0 & 1 & 0 & 1 & 0 & 1 & 0 & 1  \\
		\cmidrule(l){2-3} \cmidrule(l){4-5} \cmidrule(l){6-7} \cmidrule(l){8-9} \cmidrule(l){10-11} \cmidrule(l){12-13}  
		MaxBox & \textbf{0.60} & \textbf{0.42} & 0.06 & \textbf{0.41} & \textbf{184} & 55769 & 0.23 & \textbf{0.45} & 0.24 & \textbf{0.43} & \textbf{1621} & \textbf{37627} \\ 
		PRIM & 0.42 & 0.37 & \textbf{0.18} & 0.39 & \textbf{184} & \textbf{46070} & 0.20 & 0.42 & \textbf{0.25} & 0.39 & \textbf{1621} & 42958 \\ 
		MAIRE & 0.18 & 0.41 & 0.04 & \textbf{0.41} & \textbf{184} & 68126 & 0.06 & 0.41 & 0.11 & 0.35 & \textbf{1621} & 92976 \\ 
		Anchors & 0.27 & \textbf{0.42} & 0.16 & 0.40 & 26402 & 94448 & \textbf{0.31} & 0.42 & 0.18 & 0.36 & 77818 & 129276 \\ 
		\bottomrule 
	\end{tabular}
	\endgroup
\end{table}

\subsection{Results}
\begin{figure}[t]
	\centering
	\includegraphics[width = .9\textwidth, trim = {0 .7cm 0 0}, clip]{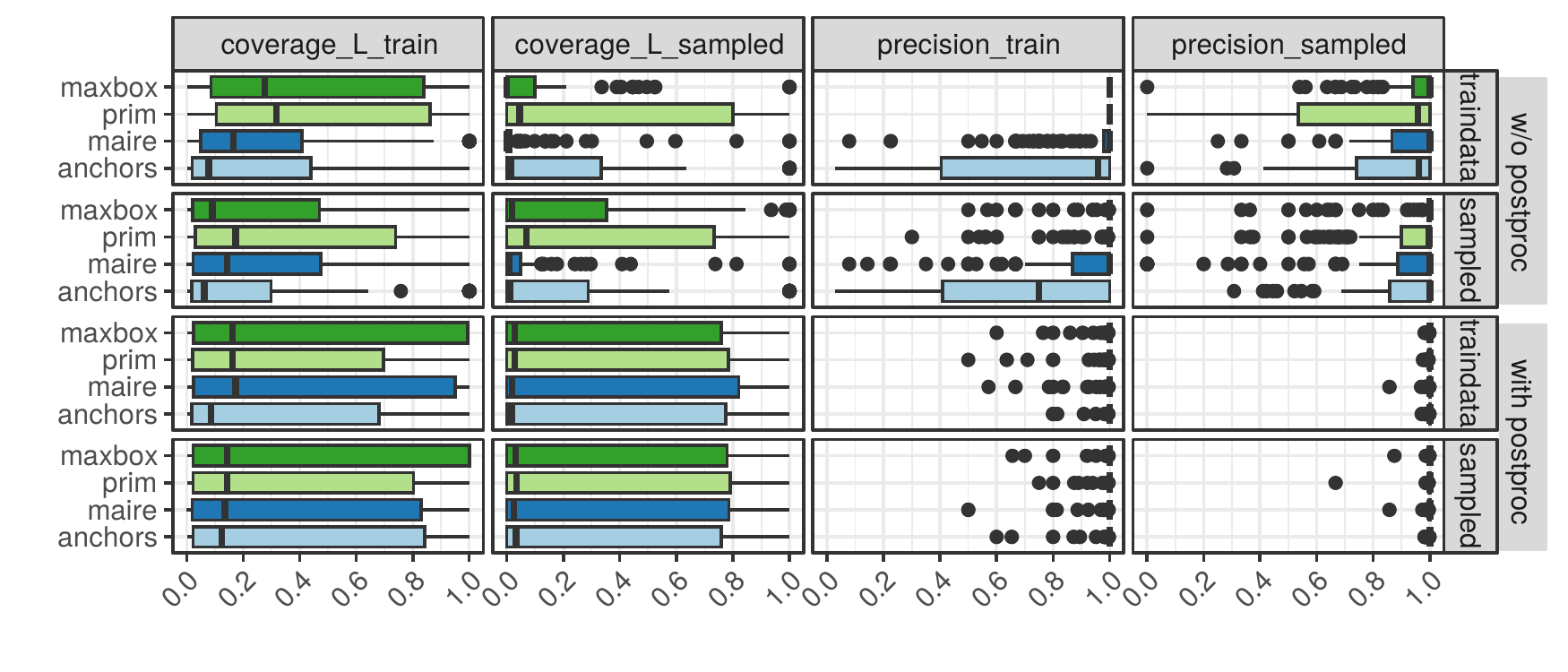}
	\caption{
		Comparison of IRD methods  w.r.t.\ coverage and precision as evaluated on the training data or newly sampled data within $\Bbar$. Addendum L means that for the coverage evaluation only training or sampled points within $\mathcal{L}$ are considered. Each point in the boxplot reflects the performance of a generated IRD of one experimental setting (dataset, model and $\xv'$). 
		Each method was either run or evaluated on training data (traindata) or uniformly sampled data from $\Bbar$ (sampled), and the methods were run either without or with post-processing (postproc). Higher values for precision and coverage are better.
	}
	\label{fig:bench}
\end{figure}

Figure~\ref{fig:bench} compares the coverage and precision values of the methods visually. Table~\ref{tab:bencheval} shows the frequency of fulfilling maximality and the number of calls to $\fh$ of the methods.
The separate results for each dataset and model, the statistical analysis, and the results of robustness are shown in Appendix~I.
We omitted the results for the locality measure because all returned IRDs covered $\xv'$.

\textit{RQ 1 - comparison of methods}
Without post-processing and training data as $\Xv$ (first row, Figure~\ref{fig:bench}), MaxBox had the highest precision as evaluated on training and newly sampled data, followed by MAIRE. 
The IRDs of PRIM had on average the largest coverage, but they also covered sampled data with predictions outside $Y'$. Due to the randomized choice of a subbox in the case of ties, PRIM is not robust according to our robustness metric. 
None of the methods outperformed the other methods w.r.t.\ maximality.
Overall, all methods outperformed the baseline method Anchors according to coverage and precision and calls to $\fh$.
The latter is because competing boxes are evaluated on column-wise permutations of the observed data. All other methods only called $\fh$ $|\Xv|$ times.

\textit{RQ 2 - training vs. sampled data}
On average, double-in-size sampled data originating from $\Bbar$ led to slightly higher coverage, precision and maximality rates w.r.t.\ newly sampled data but not w.r.t.\ the training data.
Due to the increase in the size of $\Xv$, more calls to $\fh$ were necessary.\footnote{The size decuples instead of doubles compared to the training data, because not all training data are $ \in \Bbar$ and, thus, not in $\Xv$.} 

\textit{RQ 3 - without vs. with post-processing}
Post-processing increased the coverage and precision of IRDs for all methods. 
The difference in the quality of IRDs between the methods and between the underlying data scheme (training data vs. sampled data) diminished.
Quality enhancement comes at the cost of efficiency and robustness; on average, post-processing resulted in 57,000 additional calls to $\fh$ and the sampling of new data decreased the robustness. 
MAIRE required on average the most post-processing iterations, followed by Anchors.


\section{Conclusion, Limitations and Outlook}
\label{sec:conclusion}

\paragraph{Conclusion} We introduced IRDs that describe regions in the feature space that do not affect the prediction of an instance in the form of hyperboxes.
These hyperboxes provide a set of semi-factual explanations to justify a prediction, and indicate which features affect a prediction and whether there might be pointwise biases or implausibilities. 
We formalized the search for IRDs, and introduced desiderata, a unifying framework and quality measures for IRD methods.
We discussed three existing hyperbox methods in detail and adapted them to search for IRDs.
The lack of a method ``ruling it all'' in the benchmark study emphasizes the need for a unifying framework comprising multiple methods.
The study also revealed that access to a larger, uniformly sampled dataset or using our proposed post-processing method can further enhance the quality of IRDs. 

\paragraph{Limitations} Our work offers potential for further research, e.g., on the sensitivity of the methods' hyperparameters, on the influence of sampling sizes, or on the methods' robustness w.r.t.\ slight changes in $\xv'$ or the underlying data.
While we only considered low-dimensional datasets in the benchmark study, for high-dimensional datasets we proposed two strategies to restrict the search space: 
either by letting users decide which features can be changed and which cannot (Section~\ref{subsec:desiderata}), or by deriving the largest local box $B \subset \Bbar$ based on ICE curves (Section~\ref{subsec:searchspace}).
Further research can explore: (1) the use of other IML methods, such as feature importance methods, to select features for which changes are investigated (all other features are set to their admissible value range); (2) the consideration of feature correlations or causal relations to generate IRDs, which not only naturally restricts the search space but also makes the IRD faithful to the DGP.
Considering feature correlations is also important for the application of IRDs beyond tabular data. While all presented methods are model-agnostic, we leave concrete investigations on image and text data to future research.

\paragraph{Outlook} We believe that our work can also be a starting point for investigations on the application of IRDs in other fields, e.g., for hyperparameter (HP) tuning: if a promising HP set for an ML model was identified by a tuning method, IRDs can reveal its sensitivity and whether there are other equally good but more efficient HP settings. 
IRDs might also identify high-fidelity regions for interpretable local surrogate models, like LIME~\cite{ribeiro_lime_2016}. LIME approximates predictions of a black-box model $\fh(\xv)$ around an observation $\xv'$ using a (regularized) linear model $\hat{g}(\xv)$. Here, it might be useful to understand in which region $B$ the linear model approximates the black-box model (high-fidelity region);
$\hat{g}$ only provides valuable insights in the region $B$ around $\xv'$ where $\forall \xv \in B: \hat{h}(\xv) := |\fh(\xv) - \hat{g}(\xv)| \le \epsilon$ for a user-defined $\epsilon > 0$. With $\hat{h}$ as the prediction model and $Y' = [0, \epsilon]$, IRD methods might identify such high-fidelity regions $B$ in an interpretable manner.

\section*{Acknowledgements}
This work has been partially supported by the Federal Statistical Office of Germany.





\bibliography{literature.bib}
\bibliographystyle{splncs04}



\appendix



\section{Application Examples}
\label{ap:applications}

In addition to the credit application in Section~\ref{sec:intro}, we show in the following a medical and jurisdictional application.

\paragraph{Medical} Consider an ML model that predicts if a person will develop diabetes in the future. (For simplicity, we assume this model accurately approximates real world relationships.) In the following, we discuss two cases:

(1) A person that is predicted to develop diabetes wants to know
why this is the case and what can be options to prevent this. There are different potential actions to take: more sport, less red meat, homeopathic medicine, etc. The IRD can tell which action is not promising, e.g., sports when all realistic amounts of sport are inside the box. However, changing the diet might be an option, because changing the diet by just eating meat one day a week is not part of the box (concrete strategies for prevention can reveal counterfactual explanations).

(2) A person that is predicted not to develop diabetes wants to know how flexible their life-style is without changing the prediction. It may be okay for a person to gain weight without having a higher risk of developing diabetes, as long as they do not change their diet towards including more red meat.

\paragraph{Jurisdiction} Consider an ML model that predicts if a person will commit a crime in the next 2 years. A person that gets a high score wants to know why. IRDs that do not contain all groups of protected attributes, such as gender, can indicate unfair discrimination against these groups. Hence, IRDs can initiate further investigations on fairness and biases of an ML model.

\section{Proof of Theorem \ref{theorem:sparsity}}
\label{ap:sparsity}



\begin{proof}
	Given a feature $X_j$ that is not involved in the prediction model $\fh$ such that $\forall \tilde{\xv} \in \Xspace \land \forall x_j \in \Xspace_j$:
	\begin{equation}
	\fh(\tilde{x}_1, ..., \tilde{x}_{j-1}, \tilde{x}_j, \tilde{x}_{j+1}, ..., \tilde{x}_p) = \fh(\tilde{x}_1, ..., \tilde{x}_{j-1}, x_j, \tilde{x}_{j+1}, ..., \tilde{x}_p),
	\label{eq:notinvolved}
	\end{equation}
	and given a box $B$ for $\xv'$ that is maximal according to Definition~\ref{def:maximality}.
	We assume now that Theorem~\ref{theorem:sparsity} does not hold such that $B_j = [l_j, u_j] \subset \Xspace_j$.
	However, since Eq.~(\ref{eq:notinvolved}) holds, either $(\exists\, x_j \in \Xspace_j \wedge x_j < l_j: precision(B \cup [x_j, l_j]) = 1)$, or $ (\exists\, x_j \in \Xspace_j \wedge x_j > u_j: precision(B \cup [u_j, x_j]) = 1)$ for numeric $X_j$ or $(\exists\, x_j \in \Xspace_j\setminus B_j : precision(B \cup x_j) = 1)$ for categorical $X_j$ holds which contradicts the maximality assumption of $B$.
\end{proof}

\section{Proof of Theorem \ref{theorem:detectionpower}}
\label{ap:detectionpower}


\begin{proof}
	
	Given a box $B$ with $precision(B) = 1$ and $\xv' \in B$, and given a feature $X_j$ that is relevant for $\fh(x')$ such that 
	$\exists x_j \in \Xspace_j \setminus B_j: \fh(x'_1, ..., x'_{j-1}, x_j, x'_{j+1}, ..., x'_p) \not \in Y'$. 
	We assume now that Theorem~\ref{theorem:detectionpower} does not hold, such that $B_j = \Xspace_j$. 
	This contradicts the statement that $precision(B) = 1$ because $x_j$ that leads to a prediction $\not \in Y'$ for $\xv'$ is also covered by the box.
	
\end{proof}

\section{Proof of Theorem \ref{theorem:barBbar}}
\label{ap:barBbar}

\begin{proof}
	Without loss of generality, we assume that we only have numeric features.
	Assume we computed $\Bbar = \bigcup_{j = 1}^p [l_j, u_j]$ such that 
	$\forall j \in \{1, ... p\}:$
	$$ \fh(\underbrace{x'_1, .., x'_{j-1}, l_j, x'_{j+1}, ..., x'_p}_{:=\xv'_l}) \not \in Y' \land \fh(\underbrace{x'_1, .., x'_{j-1}, u_j, x'_{j+1}, ..., x'_p}_{:=\xv'_u}) \not \in Y'.$$  
	We assume that $B \subset \Bbar$ is not true for now such that there is a homogeneous $B$ with $min(B_j) < l_j$ or $max(B_j) > u_j$ and $\xv' \in B$.
	However, then either $\xv'_l$ or $\xv'_u$ would also be part of $B$ but for both $\fh(\xv'_u) \not \in Y'$ or $\fh(\xv'_l) \not \in Y'$ holds, which contradicts that $B$ is homogeneous.  
\end{proof}

\clearpage

\section{Pseudocode and Illustrations of IRD Methods}
\label{ap:methods}

\subsection{Pseudocode}

\begin{algorithm}[h]
	\caption{Adapted MaxBox approach \cite{eckstein_maxbox_2002}}
	\label{algo:maxbox}
	\begin{algorithmic}
		\STATE {\bfseries Input:} Targeted instance $\xv'$, desired range $Y'$, prediction model $\fh: \mathcal{X} \rightarrow \mathbb{R}$, 
		input dataset $\Xv$, initial box $B$
		\STATE {Initialize candidates $= [\,]$, upper\_bound\_coverage\_best = -Inf, current\_best $= [\,]$}
		\IF{$\exists \xv \in \Xv \land \xv \in B: \fh \not \in Y'$}
		\STATE {candidates = candidates.$append(B)$}
		\WHILE {$length$(candidates) $> 0$}
		\STATE {$B^{best} = choose\_best($candidates)}  
		\STATE \hspace{0.5cm} {$\triangleright$ \parbox[t]{.9\linewidth}{if upper\_bound\_coverage\_best $< 0$, $B^{best}$ corresponds to the box with the most no.\ of shrinking steps done before (with the upper bound of the coverage as a tiebreaker), else, $B^{best}$ corresponds to the box that maximizes  $\left(\frac{|\{\xv \in B| \fh(\xv) \in Y\}|}{|\{\xv \in B| \fh(\xv) \not \in Y\}|}\right)$.}}
		\STATE candidates = candidates.$remove(B^{best})$
		\STATE children = $create\_new\_candidates(B^{best})$ \hspace{0cm} {$\triangleright$ in Figure \ref{fig:maxbox}, C and D are new candidates created from the initial box}
		\FOR{$B \in$ children}
		\IF{$\forall \xv \in B: \fh(\xv) \in Y'$}
		\STATE {coverage $= upper\_bound\_coverage(B)$}
		\IF{coverage $>$ upper\_bound\_coverage\_best}
		\STATE {current\_best $= B$}
		\STATE {upper\_bound\_coverage\_best = coverage}
		\ENDIF
		\ELSE 
		\IF {$upper\_bound\_coverage(B) >$ upper\_bound\_coverage\_best} 
		\STATE {candidates = candidates.$append(B)$}
		\ENDIF
		\ENDIF
		\ENDFOR
		\ENDWHILE
		\ELSE
		\STATE {current\_best = B}
		\ENDIF
		\STATE {\bfseries return} current\_best
	\end{algorithmic}
\end{algorithm}

\clearpage

\vspace{-.7cm}
\begin{algorithm}[H]
	\caption{Adapted PRIM approach \cite{friedman_bump_1999}}
	\label{algo:prim}
	\begin{algorithmic}
		\STATE {\bfseries Input:} Targeted instance $\xv'$, desired range $Y'$, prediction model $\fh: \mathcal{X} \rightarrow \mathbb{R}$, 
		input dataset $\Xv$, initial box $B$
		\WHILE {$\exists \xv \in \Xv \land \xv \in B: \fh \not \in Y'$}
		\FOR{$j \in \{1, ..., p\}$}
		\STATE {$C_j = [\,]$} \hspace{3cm} {$\triangleright$ create candidates for peeling}
		\IF {$X_j$ numeric} 
		\STATE {$C_j = C_j.append(B_j^-, B_j^+)$ where $B_j^- = [l_j, min(X_{j(\alpha)}, x_j')]$ and \\$B_j^+ = [max(X_{j(1-\alpha)}, x_j'), u_j]$ 
			with $x_{j(\alpha)}$ and $x_{j(1-\alpha)}$ as the $\alpha$- and $(1-\alpha)$-quantiles of $X_j$ in the current box $B$}
		\ELSIF {$X_j$ categorical}
		\STATE $C_j = \{s \in B_j \mid s \neq x'_j\}$
		\ENDIF
		\ENDFOR
		\STATE $b^{best} = \underset{b \in C_j,\, j \in \{1, ..., p\}}{\arg \max} \, precision(B \setminus b)$
		\STATE $B = B \setminus b^{best}$
		\ENDWHILE
		\STATE homogeneous = TRUE
		\WHILE {homogeneous}
		\FOR{$j \in \{1, ..., p\}$}
		\STATE {$C_j = [\,]$} \hspace{5cm} {$\triangleright$ create candidates for pasting}
		\IF {$X_j$ numeric} 
		\STATE inbox = $\{\xv \in \Xv \mid x_k \in B_k\}$, for $k \in \{1, ..., j-1, j+1, ... p\}$
		\STATE number\_added $ = |\{\xv \in \Xv \mid \xv \in B\}| \cdot \alpha$
		\STATE $C_j = C_j.append(B_j^-, B_j^+)$ with $B_j^- = [x_j^l, l_j]$ and $B_j^+ = [u_j, x_j^u]$ with \\
		$x_j^l$ as the $j$th feature value of the (number\_added)th observation $\xv \in$ inbox with a value $x_j$ lower than $l_j$ and \\
		$x_j^u$ as the $j$th feature value of the (number\_added)th observation $\xv \in$ inbox with a value $x_j$ higher than $u_j$ \\
		\ELSIF {$X_j$ categorical}
		\STATE $C_j = \{s \in X_j \mid s \not \in B_j \}$
		\ENDIF
		\STATE $C_j = \{b \in C_j \mid precision(B \cup b) = 1\}$
		\ENDFOR  
		\IF {$\exists j \in \{1,..., p\}: |C_j| > 0$}
		\STATE $b^{best} = \underset{b \in C_j,\, j \in \{1, ..., p\}}{\arg \max} \, \, coverage(B \setminus b)$
		\STATE $B = B \cup b$
		\ELSE 
		\STATE homogeneous = FALSE
		\ENDIF
		\ENDWHILE
		\STATE {\bfseries return} B
	\end{algorithmic}
\end{algorithm}


\begin{algorithm}[H]
	\caption{Adapted MAIRE approach \cite{sharma_maire_2021}}
	\label{algo:maire}
	\begin{algorithmic}
		\STATE {\bfseries Input:} Targeted instance $\xv'$, desired range $Y'$, prediction model $\fh: \mathcal{X} \rightarrow \mathbb{R}$, 
		input dataset $\Xv$, initial box $B$, 
		precision threshold $\tau$ (default 1), 
		maximum number of iterations max\_iterations (default 100)   
		\STATE Scale all feature values of $\xv \in \Xv$ and $\xv'$ to 0-1 range
		\STATE best\_coverage $= 0$
		\STATE converged = FALSE
		\STATE best\_candidate = $B$
		\STATE $i = 0$
		\WHILE {$i \le max\_iterations$}
		\STATE $B = optimize\_with\_adam(B)$  
		\STATE \hspace{.5cm} {$\triangleright$ optimizes differentiable versions of coverage, precision and locality}
		\IF {$precision(B) \ge \tau \land coverage(B) \ge$ best\_coverage}
		\STATE  best\_candidate $= B$
		\ELSIF {$precision(B) < \tau$}
		\STATE converged = TRUE
		\ENDIF
		\IF {converged = TRUE}
		\STATE i = i + 1
		\ENDIF
		\ENDWHILE
		\STATE {\bfseries return} best\_candidate
	\end{algorithmic}
\end{algorithm}




\subsection{Illustrations}

\begin{figure*}[h!]
	\includegraphics[width=1\textwidth, page = 2]{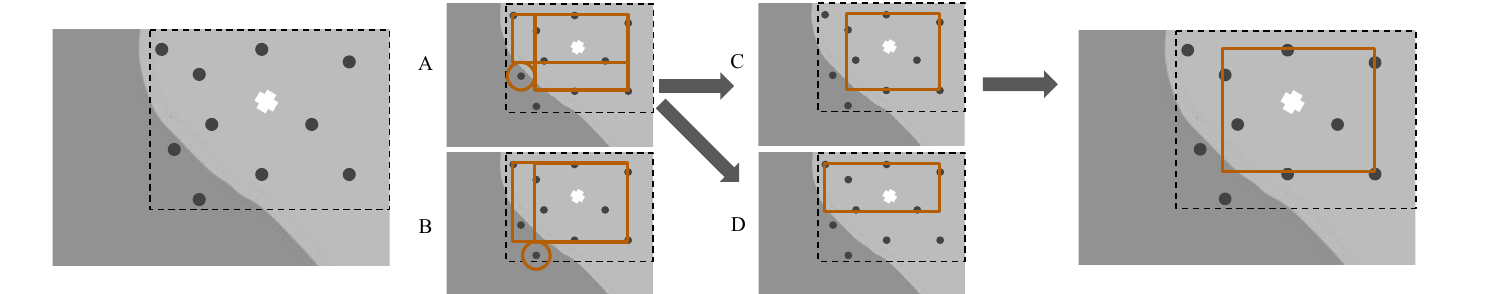}
	\caption{Illustration of the adapted MaxBox algorithm. The algorithm starts with $\Bbar$ (dashed box). In the box are two data points with predictions $\not \in Y'$ (called negative samples) and the box needs to be further optimized. 
		First, a negative sample is chosen - either the one in A or B. Therefore, the number of samples with predictions $\in Y'$ after excluding the points in one feature dimension are inspected.  The resulting boxes of both negative samples cover a maximum of seven samples. We chose the one of A (B is also fine). Its resulting boxes are the new subproblems/candidates (C and D). Both boxes in C and D only include samples with predictions $\in Y'$, but the box in C is chosen as an optimum because it includes more samples with predictions $\in Y'$. D is discarded because it has a lower number. Since C and D cannot be further split because no negative samples are within both boxes, the returned box by MaxBox is the box in C.}
	\label{fig:maxbox}
\end{figure*}

\begin{figure*}[h!]
	\includegraphics[width=1\textwidth, page = 4]{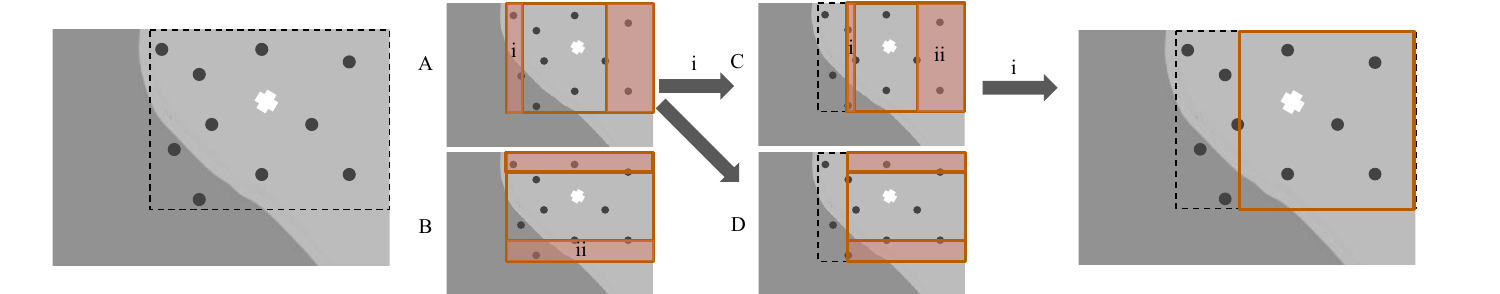}
	\label{fig:prim}
	\caption{Illustration of the adapted PRIM algorithm. The algorithm starts with $\Bbar$. In the first iteration, there exist four potential subboxes (two in each feature dimension (A vs. B)) that could be removed. The subbox i is chosen because it has the highest precision but compared to ii it has a smaller size. In the next step (C \& D), again four subboxes can be potentially removed. Again, we choose i for the same reason as before. After its removal, the resulting box is at the same time the final box because in the pasting step only one subbox could be added -- i again. All other dimensions are maximal.}
\end{figure*}


\begin{figure*}[ht]
	\includegraphics[width=1\textwidth, page = 5]{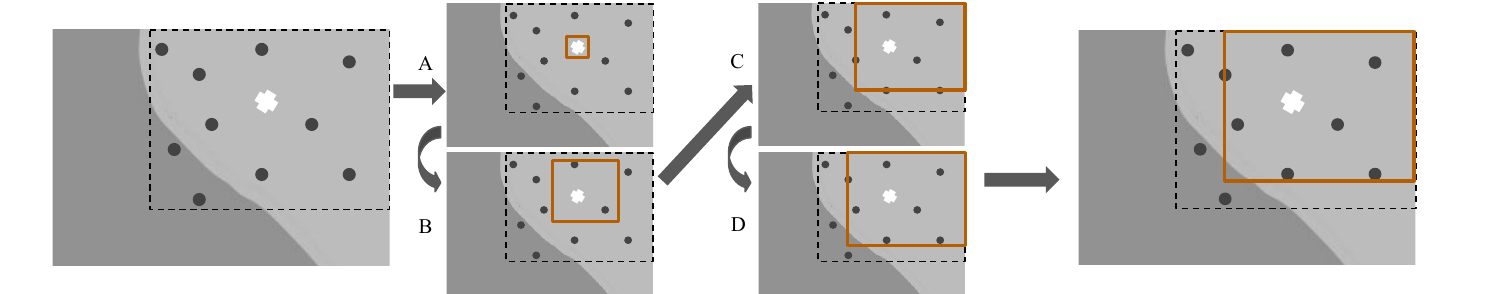}
	\label{fig:maire}
	\caption{Illustration of the adapted MAIRE algorithm. The algorithm starts with the smallest box possible. The box boundaries are then iteratively enlarged (A-D). The box boundaries are only updated if the precision of the new box $= 1$.}
\end{figure*}

\clearpage 

\section{Pseudocode of post-processing Approach}
\label{ap:postprocessing}
\vspace{-0.5cm}
\begin{algorithm}[H]
	\caption{Post-processing algorithm - peeling (inspired by \cite{friedman_bump_1999})}
	\label{algo:postproc}
	\begin{algorithmic}
		\STATE {\bfseries Input:} Targeted instance $\xv'$, desired range $Y'$, prediction model $\fh: \mathcal{X} \rightarrow \mathbb{R}$, initial box $B$, number of samples for evaluation $M$ (default 100),  relative subbox size of continuous features $\alpha$ (default $0.1$)
		\FOR{$j \in \{1, ..., p\}$}
		\IF {$X_j$ numeric} 
		\STATE $s_j = (max(\Xspace_j) - min(\Xspace_j))\cdot \alpha$ \hspace{2cm}  {$\triangleright$ derive subbox sizes for numeric features based on $\Xspace$}
		\IF {$X_j$ integer}
		\STATE $s_j = round(s_j)$
		\ENDIF
		\ENDIF
		\ENDFOR
		\STATE $\Xv = sample\_uniformly(B, n = M \cdot 5)$   \hspace{2cm}  {$\triangleright$ sample new data to check if $B$ homogeneous}
		\IF{$\exists \xv \in \Xv \land \xv \in B: \fh \not \in Y'$}
		\STATE not\_homogeneous = TRUE \hspace{3cm} {$\triangleright$ start peeling}
		\WHILE {not\_homogeneous}
		\FOR{$j \in \{1, ..., p\}$} 
		\STATE {$C_j = [\,]$} \hspace{3cm} {$\triangleright$ create candidates for peeling}
		\IF {$X_j$ numeric} 
		\STATE $C_j = C_j.append(B_j^-, B_j^+)$ \\ where $B_j^- = [l_j, min(l_j + s_j, x'_j)]$ and $B_j^+ = [max(u_j-s_j, x_j'), u_j]$
		\ELSIF {$X_j$ categorical}
		\STATE $C_j = \{s \in B_j \mid s \neq x'_j\}$
		\ENDIF
		\STATE $C_j = \{b \in C_j \mid precision(B^b_j) < 1\}$
		with $B^b_j = (B_1 \times ... \times B_{j-1} \times b \times B_{j+1} \times ...  \times B_p)$
		\ENDFOR  
		\IF {$\exists j \in \{1,..., p\}: |C_j| > 0$}
		\STATE {$b^{best} = \underset{b \in C_j,\, j \in \{1, ..., p\}}{\arg \max} \, precision\_to\_boxsize(B^b_j)$ \hspace{.3cm} {$\triangleright$ evaluate on $M$ new instances sampled within $B^b_j$}}
		\STATE $B^{best} = (B_1 \times ... \times B_{j-1} \times b^{best} \times B_{j+1} \times ...  \times B_p)$  {$\triangleright$ choose the one with lowest precision relative to size}
		\STATE $B = B^{best}$
		\ELSE 
		\STATE not\_homogeneous = FALSE
		\ENDIF
		\ENDWHILE
		\ENDIF
		\STATE {\bfseries return} B, $\mathbf{s} = \{s_j \mid X_j \text{ numeric}\}$
	\end{algorithmic}
\end{algorithm}

\begin{algorithm}[h]
	\caption{Post-processing algorithm - pasting (inspired by \cite{friedman_bump_1999})}
	\label{algo:postprocpasting}
	\begin{algorithmic}
		\STATE {\bfseries Input:} Targeted instance $\xv'$, desired range $Y'$, prediction model $\fh: \mathcal{X} \rightarrow \mathbb{R}$, initial box $B$ (potentially peeled), number of samples for evaluation $M$ (default 100),  relative subbox size of continuous features $\alpha$ (default $0.1$), lower threshold for relative subbox size $\alpha_0$ (default $0.05$), subbox sizes of numeric features $\mathbf{s}$
		
		\STATE homogeneous = TRUE \hspace{3cm} {$\triangleright$ start pasting}
		\STATE stepsize $= 1$
		\WHILE {homogeneous}
		\FOR{$j \in \{1, ..., p\}$}
		\STATE {$C_j = [\,]$} \hspace{3cm} {$\triangleright$ create candidates/subboxes for pasting}
		\IF {$X_j$ numeric} 
		\STATE $C_j = C_j.append(B_j^-, B_j^+)$  \\ 
		where $B_j^- = [l_j - \text{stepsize} \cdot s_j, l_j]$ and $B_j^+ = [u_j, u_j + \text{stepsize} \cdot s_j]$
		\ELSIF {$X_j$ categorical}
		\STATE $C_j = \{s \in X_j \mid s \not \in B_j \}$
		\ENDIF
		\STATE $C_j = \{b \in C_j \mid precision(B_j^b) = 1\}$ with $B^b_j = (B_1 \times ... \times B_{j-1} \times b \times B_{j+1} \times ...  \times B_p)$ 
		\ENDFOR  
		\IF {$\exists j \in \{1,..., p\}: |C_j| > 0$}
		\STATE $b^{best} = \underset{b \in C_j,\, j \in \{1, ..., p\}}{\arg \max} \, size(B_j^b)$ \hspace{.3cm} {$\triangleright$ evaluate on $M$ new instances sampled within  $B_j^b$}
		\STATE $B = B \cup b$ \hspace{.3cm} {$\triangleright$ choose largest one with precision 1}
		\ELSE 
		\IF{stepsize $\ge \alpha_0$}
		\STATE stepsize = stepsize$/2$  \hspace{3cm} {$\triangleright$ if no box with precision 1 exists, consider reducing the subbox sizes}
		\ELSE 
		\STATE homogeneous = FALSE
		\ENDIF
		\ENDIF
		\ENDWHILE
		\STATE {\bfseries return} B
	\end{algorithmic}
\end{algorithm}

\clearpage

\begin{figure*}[t]
	\caption{Illustration of the post-processing algorithm. The algorithm starts with the box generated by another method (solid brown box, which is a subbox of the dashed box $\Bbar$). First, new points are sampled and it is assessed whether the box is homogeneous (A). If not, the subboxes with the lowest precision compared to their size are peeled iteratively (B). The precision is assessed based on newly sampled points within the subboxes. First subbox i is peeled then subbox ii (both contain a sample with a prediction $\not \in Y'$).  If no subbox with precision $<$ 1 exists, it is assessed whether the box could be further enlarged (C). If all considered subboxes have precisions $<$ 1, the subbox sizes are halved (D) as long as the relative subbox size does not fall below a threshold.}
	\includegraphics[width=1\textwidth, page = 6]{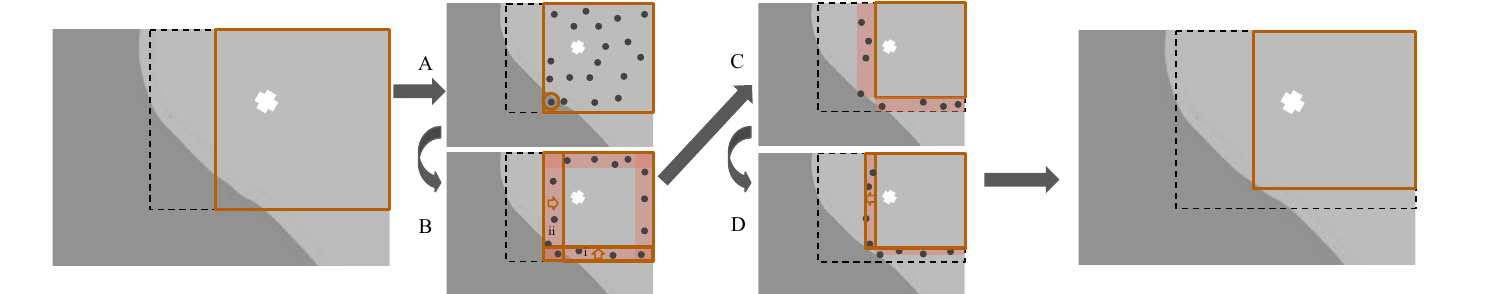}
	\label{fig:postprocessing}
\end{figure*}

\section{Level Set Identification}
\label{ap:levelset}

The algorithm starts at $\xv'$ and tries to find a connection $\in Y'$ between the nominal, then the ordinal, and then the continuous features of $\xv$ and $\xv'$.
If a path is found, $\xv$ is part of $\mathcal{L}$. 
For categorical features, all permutations of feature orders are inspected.\footnote{If the number of permutations exceeds 100 permutations, 100 feature orders are randomly chosen.} 
For continuous features, the shortest linear path for a given number of equidistant steps is checked. Kuratomi et al.\ \cite{kuratomi_juice_2022} used DBSCAN, for which the choice of the maximum distance threshold is ambiguous.
The identification algorithm has a complexity of $O(c! \cdot c + o! \cdot \sum_{j = 1}^o k_j + q)$ with $c$ and $o$ as the number of nominal and ordinal features, respectively, $k_j$ as the number of possible values of an ordinal feature $X_j$ and $q$ as the number of inspected steps for continuous features.

The level set could be further enriched by attempting to find connections between the unconnected and connected points. For the comparison of IRD methods, however, a convex level set is sufficient, since the hyperbox itself is convex.



\section{Tuning of ML models}
\label{ap:tuning}

For hyperparameter tuning, we used random search (with 15 evaluations), and 5-fold cross-validation (CV) with the
misclassification error (classification) or mean squared error (regression) as a performance measure. Table~\ref{tab:hyperparams} shows the tuning search space of each
model. 
The rather limited tuning setup should be sufficient for our task of explaining a prediction model -- a less accurate model is not a hindrance.
Unbalanced datasets such as \textit{tic\_tac\_toe}, \textit{diabetes} and \textit{cmc} were balanced with the SMOTE algorithm \cite{chawala_etal_smote_2002}. For SMOTE,
numeric features were standardized and categorical ones were one-hot encoded.
The optimizer for the neural network was ADAM \cite{kigma_2017_adam} with 500 epochs.
For all other hyperparameters, the default values of the mlr3keras R package were used \cite{pfisterer_etal_mlr3keras_2022} (apart from the no.\ of layer units, see Table~\ref{tab:hyperparams}).
Table~\ref{tab:accuracies} shows the accuracies of each model using nested resampling with 5-fold CV in the inner and outer loop).

\begin{table}[h]
	\centering
	\caption{\label{tab:hyperparams}Tuning search space of each model. Hyperparameter values of \textit{num.trees}  were log-transformed.}
	\begin{tabular}{lll}
		\hline
		Model & Hyperparameter & Range\\ \hline
		random forest & num.trees & {}[1, 1000]\\
		logistic regression & - & -\\
		linear model & - & - \\
		multi-nomial model & - &  - \\
		hyperbox/rpart & - &  - \\
		neural net & layer\_units & {}[1, 20]\\ \hline
	\end{tabular}
\end{table}

\begin{table}[ht]
	\caption{\label{tab:accuracies} Classification error or mean squared error (regression) of each model on each dataset. The performances were computed using nested resampling with 5-fold CV in the inner and outer loop. We did not measure the performance of the (terminal node) hyperbox model because the model differs for each $\xv'$.}
	\centering
	\begin{tabular}{lllll}
		\hline
		& Random forest & Linear model & Neural net & Hyperbox \\ 
		\hline
		diabetes & 0.233 & 0.224 & 0.229 & - \\ 
		tic\_tac\_toe & 0.036 & 0.019 & 0.094 & - \\ 
		cmc & 0.466 & 0.495 & 0.389 & - \\ 
		vehicle & 0.256 & 0.201 & 0.254 & - \\ 
		no2 & 33502.856 & 37678.319 & 77866.331 & - \\ 
		plasma\_retinol & 45391.218 & 59224.452 & 297481.249 & - \\ 
		\hline
	\end{tabular}
\end{table}


\begin{figure}[h]
	\centering
	\includegraphics[width = 1\textwidth, clip, trim=1.5cm 1cm 12cm .5cm, page = 1]{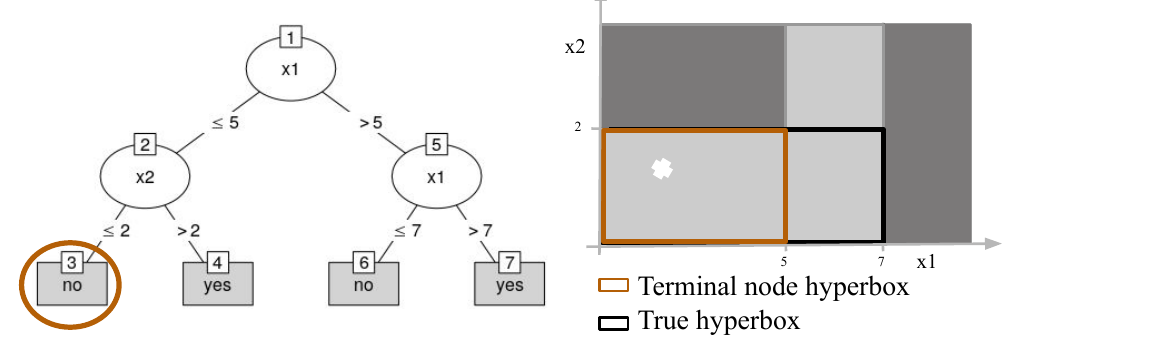}
	\caption{True hyperbox vs. terminal node hyperbox for a CART tree. The white cross corresponds to $\xv'$.}
	\label{fig:carthyperbox}
\end{figure}

\section{Benchmark - Additional Results}
\label{ap:benchadditional}

\begin{figure}[H]
	\centering
	\includegraphics[width = .9\textwidth, trim = {0 .5cm 0 0}, clip]{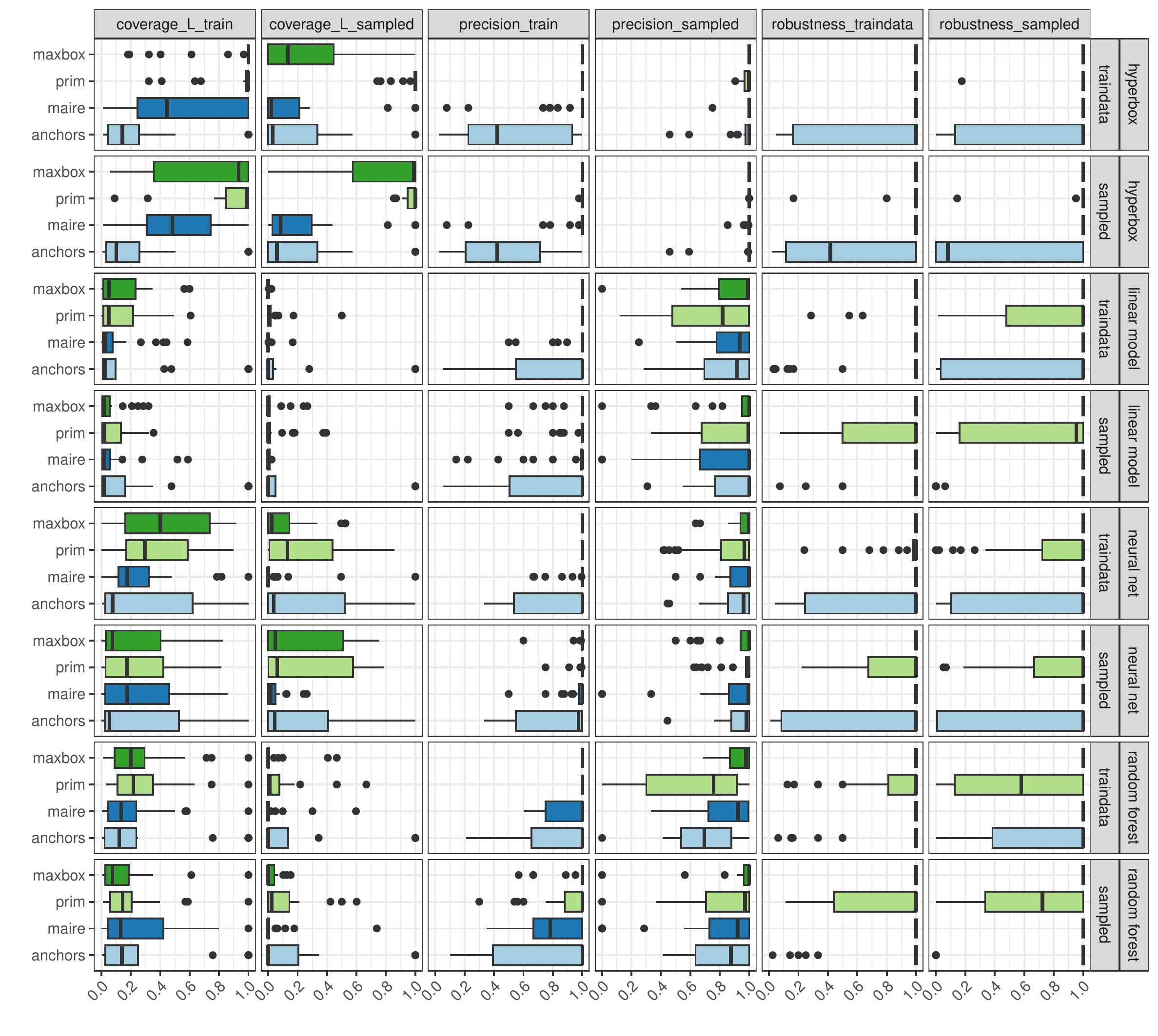}
	\caption{Comparison of MaxBox, PRIM, Anchors, and MAIRE  w.r.t.\ coverage and precision for each model separately. 
		Each method was either run or evaluated on training data (traindata) or uniformly sampled data from $\Bbar$ (sampled) \textit{without} post-processing. Higher values for precision and coverage are better.}
	\label{fig:model0}
\end{figure}

\begin{figure}[h]
	\centering
	\includegraphics[width = .9\textwidth, trim = {0 .5cm 0 0}, clip]{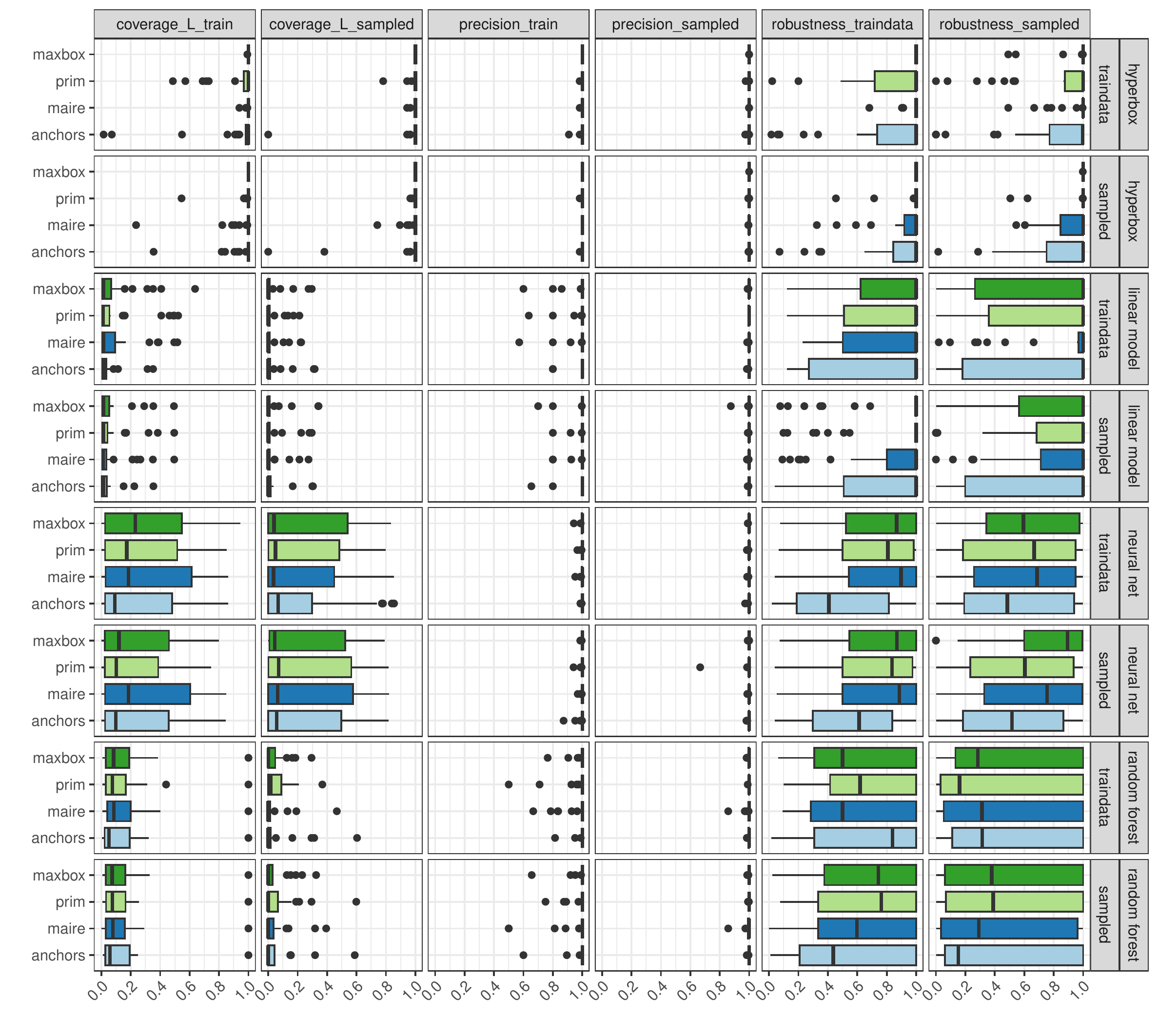}
	\caption{Comparison of MaxBox, PRIM, Anchors, and MAIRE  w.r.t.\ coverage and precision for each model separately. 
		Each method was either run or evaluated on training data (traindata) or uniformly sampled data from $\Bbar$ (sampled) \textit{with} post-processing. Higher values for precision and coverage are better.}
	\label{fig:model1}
\end{figure}

\begin{figure}[h]
	\centering
	\includegraphics[width = .9\textwidth, trim = {0 .5cm 0 0}, clip]{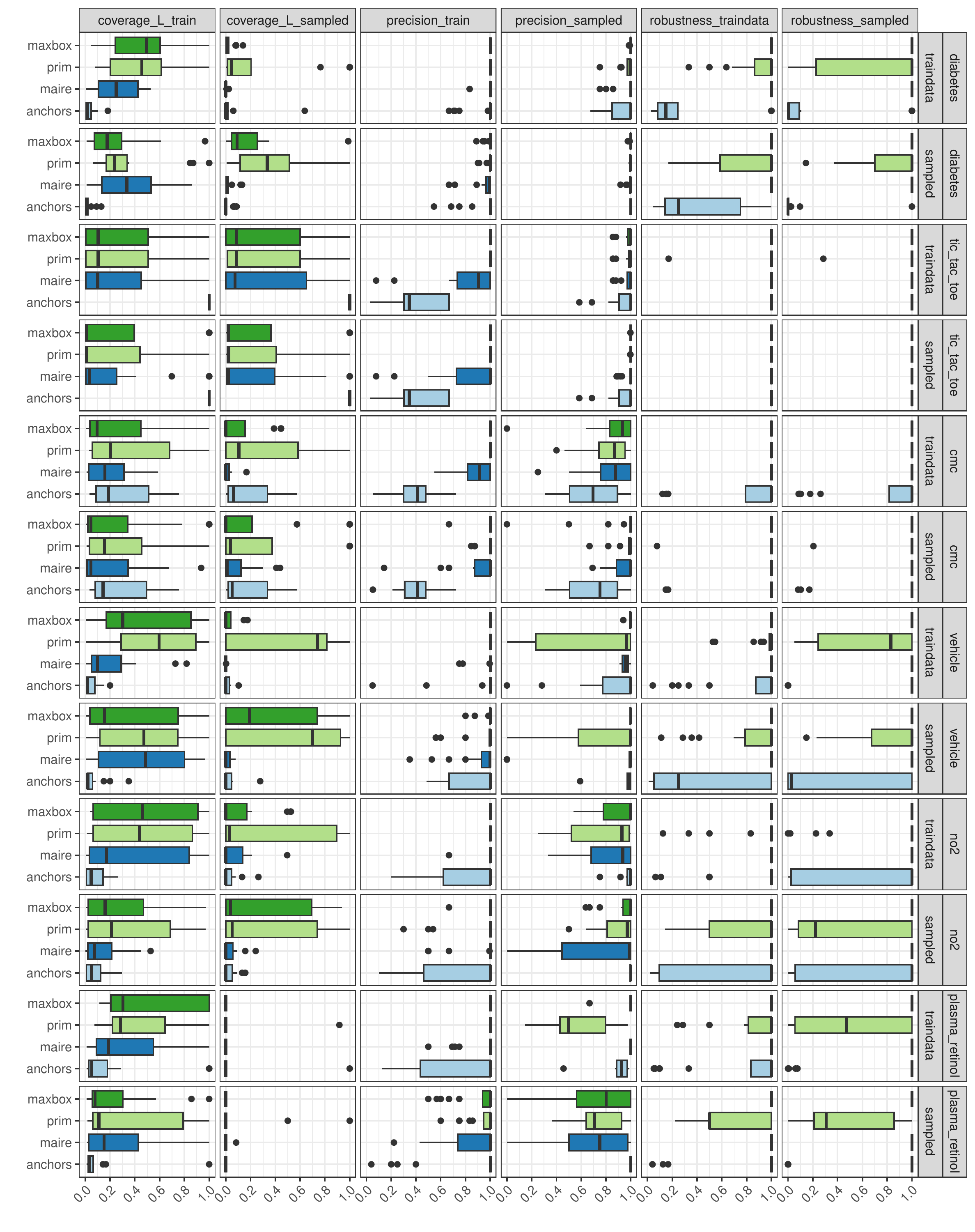}
	\caption{Comparison of MaxBox, PRIM, Anchors, and MAIRE  w.r.t.\ coverage and precision for each dataset separately. 
		Each method was either run or evaluated on training data (traindata) or uniformly sampled data from $\Bbar$ (sampled) \textit{without} post-processing. Higher values for precision and coverage are better.}
	\label{fig:dataset0}
\end{figure}

\begin{figure}[h]
	\centering
	\includegraphics[width = .9\textwidth, trim = {0 .5cm 0 0}, clip]{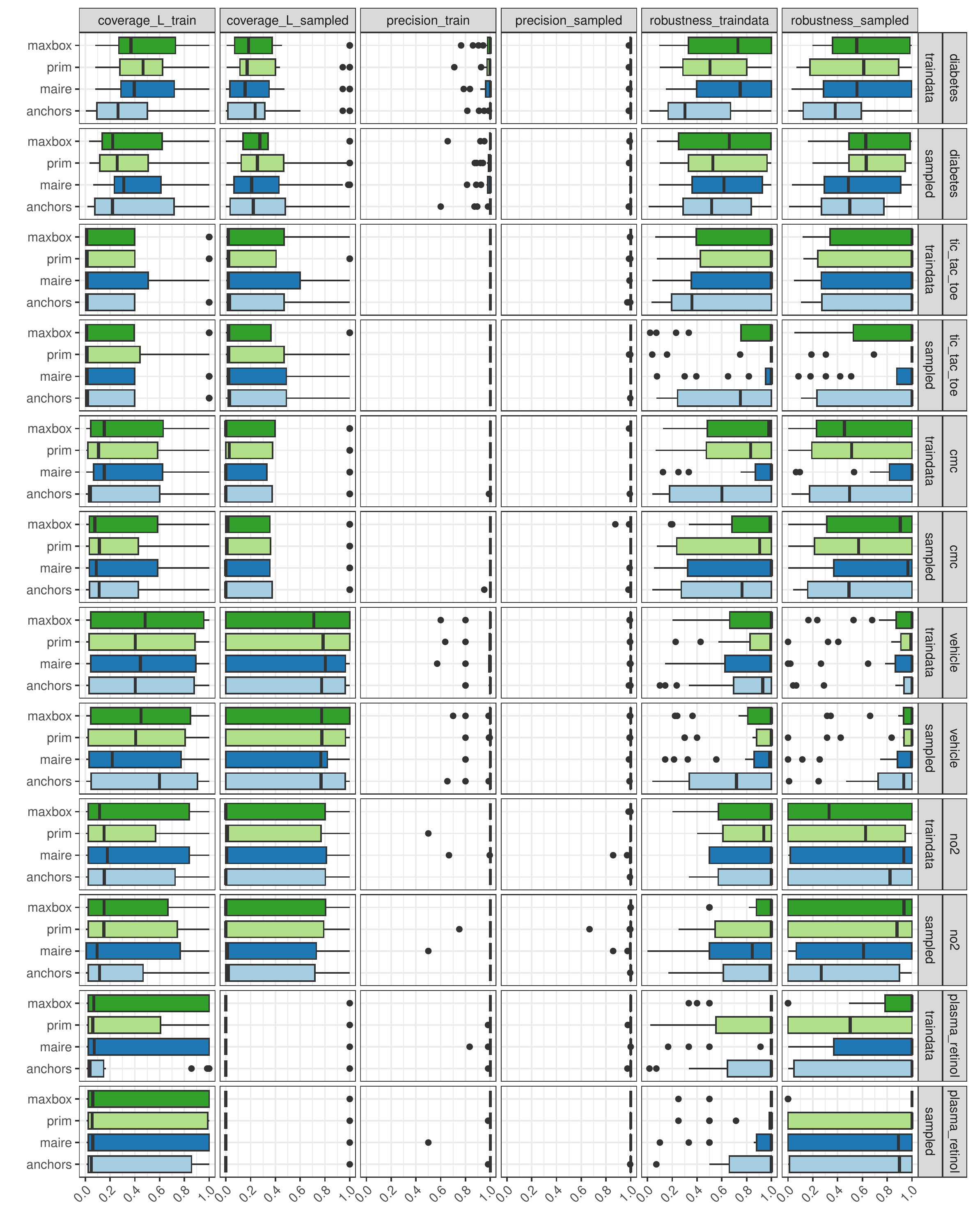}
	\caption{Comparison of MaxBox, PRIM, Anchors, and MAIRE  w.r.t.\ coverage and precision for each dataset separately. 
		Each method was either run or evaluated on training data (traindata) or uniformly sampled data from $\Bbar$ (sampled) \textit{with} post-processing. Higher values for precision and coverage are better.}
	\label{fig:dataset1}
\end{figure}

\clearpage

\begin{sidewaystable}
	\centering
	\caption{\label{tab:rq1} Statistical analysis of RQ 1. Pairwise comparison of MaxBox, PRIM, Anchors, and MAIRE  w.r.t.\ coverage and precision. Each value corresponds to the p-value obtained for the Wilcoxon rank-sum test with $H_0$ that the performances do not differ. Cells printed in bold font correspond to p-values that are lower than $\alpha = 0.05/36$ (Bonferroni-adjustment) and indicate that one method outperforms the other. Only methods run on the training data without post-processing were compared.}
	\begin{tabular}{lllllll}
		\hline
		measure & MaxBox = PRIM & MaxBox = Anchors & MaxBox = MAIRE & PRIM = Anchors & PRIM = MAIRE & Anchors = MAIRE \\ 
		\hline
		coverage\_train & 0.761 & 0.618 & \textbf{0} & 0.579 & \textbf{0} & 0.473 \\ 
		coverage\_sampled & \textbf{0} & 0.044 & \textbf{0} & \textbf{0} & \textbf{0} & \textbf{0} \\ 
		coverage\_$\mathcal{L}$\_train & 0.431 & \textbf{0.001} & \textbf{0} & \textbf{0} & \textbf{0} & 0.127 \\ 
		coverage\_$\mathcal{L}$\_sampled & \textbf{0} & 0.035 & 0.004 & 0.059 & \textbf{0} & \textbf{0} \\ 
		precision\_train & 1 & \textbf{0} & \textbf{0} & \textbf{0} & \textbf{0} & \textbf{0} \\ 
		precision\_sampled & 0.025 & \textbf{0} & 0.623 & 0.042 & 0.104 & 0.004 \\ 
		\hline
	\end{tabular}
	
	\bigskip\bigskip\bigskip
	\caption{Statistical analysis of RQ 2. Pairwise comparison of using training data vs. sampled data for $\Xv$. Each value corresponds to the p-value obtained for the Wilcoxon rank-sum test with $H_0$ that the performance of methods using training data is better than the performance of methods using sampled data. Cells printed in bold font correspond to p-values that are lower than $\alpha = 0.05/30$ (Bonferroni-adjustment)  and indicate a preference towards using sampled data. Comparisons were only conducted for the methods run without post-processing.}
	\centering
	\begin{tabular}{lrllll}
		\hline
		measure & overall & MaxBox & PRIM & Anchors & MAIRE \\ 
		\hline
		coverage\_train & 1.00 & 1 & 0.999 & 0.445 & 0.744 \\ 
		coverage\_sampled & 0.00 & \textbf{0} & 0.987 & 0.402 & 0.003 \\ 
		coverage\_$\mathcal{L}$\_train & 1.00 & 1 & 1 & 0.781 & 0.896 \\ 
		coverage\_$\mathcal{L}$\_sampled & 0.00 & \textbf{0} & 0.236 & 0.476 & 0.172 \\ 
		precision\_train & 1.00 & 0.995 & 0.998 & 0.782 & 0.993 \\ 
		precision\_sampled & 0.00 & 0.011 & \textbf{0} & \textbf{0.001} & 0.381 \\ 
		\hline
	\end{tabular}
\end{sidewaystable}

\begin{sidewaystable}
	\centering
	\caption{Statistical analysis of RQ 3. Pairwise comparison of using no post-processing vs. using post-processing. Each value corresponds to the p-value obtained for the Wilcoxon rank-sum test with $H_0$ that the performance of methods using no post-processing is better than the performance of methods using post-processing. Cells printed in bold font correspond to p-values that are lower than $\alpha = 0.05/60$ (Bonferroni-adjustment) and indicate a preference towards post-processing.}
	\centering
	\begin{tabular}{lrlllll}
		\hline
		method & coverage\_train & coverage\_sampled & coverage\_$\mathcal{L}$\_train & coverage\_$\mathcal{L}$\_sampled & precision\_train & precision\_sampled \\ 
		\hline
		\textbf{traindata} & 0.95 & 0 & 0.369 & 0 & 0 & 0 \\ 
		MaxBox & 0.97 & \textbf{0} & 0.982 & \textbf{0} & 0.995 & 0.003 \\ 
		PRIM & 1.00 & 1 & 1 & 0.452 & 0.999 & \textbf{0} \\ 
		Anchors & 0.92 & 0.001 & 0.065 & 0.054 & \textbf{0} & \textbf{0} \\ 
		MAIRE & 0.10 & \textbf{0} & 0.003 & \textbf{0} & \textbf{0} & 0.001 \\ 
		\hline
		\textbf{sampled} & 0.12 & 0 & 0 & 0 & 0 & 0 \\ 
		MaxBox & 0.00 & \textbf{0} & \textbf{0} & \textbf{0} & 0.085 & 0.021 \\ 
		PRIM & 0.45 & 0.19 & 0.262 & 0.468 & 0.003 & 0.001 \\ 
		Anchors & 0.92 & \textbf{0} & 0.035 & 0.061 & \textbf{0} & \textbf{0} \\ 
		MAIRE & 0.18 & \textbf{0} & 0.003 & \textbf{0} & \textbf{0} & 0.009 \\ 
		\hline
	\end{tabular}
\end{sidewaystable}

\end{document}